\documentclass[runningheads]{llncs}
\usepackage{graphicx}
\usepackage{acronym}  
\usepackage{tikz}
\usepackage{tkz-euclide} 
\usepackage{amsmath}
\usepackage{amsfonts}
\usepackage{changepage}
\usepackage{caption}
\usepackage{subcaption}
\DeclareMathOperator*{\argmax}{argmax}
\DeclareMathOperator*{\argmin}{argmin}
\DeclareMathOperator*{\E}{\mathbb{E}}

\usepackage[ruled]{algorithm2e}
\usepackage{amstext} 
\usepackage{array}   
\newcolumntype{L}{>{$}l<{$}} 
\newcolumntype{C}{>{$}c<{$}} 

\usepackage[misc]{ifsym}

\begin{document}
\newacro{r.v.}{random variable}
\newacroplural{r.v.}[r.v.s]{random variables}
\newacro{MSE}{Mean Squared Error}
\newacro{DNN}{Deep Neural Network}
\newacroplural{DNNs}{Deep Neural Networks}
\newacro{CNN}{Convolutional Neural Network}
\newacro{ML}{Machine Learning}
\newacro{ML4H}{Machine Learning for Health}
\newacro{DL}{Deep Learning}
\newacro{AI}{Artificial Intelligence}
\newacro{MRI}{Magnetic Resonance Imaging}
\newacroplural{MRI}[MRIs]{Magnetic Resonance Images}
\newacro{GAN}{Generative Adversarial Network}
\newacroplural{GAN}[GANs]{Generative Adversarial Networks}
\newacro{VAE}{Variational Autoencoder}
\newacroplural{VAE}[VAEs]{Variational Autoencoders}
\newacro{MLE}{Maximum Likelihood Estimation}
\newacro{MI}{Mutual Information}
\newacro{PCA}{Principal Component Analysis}
\newacro{CCA}{Canonical Correlation Analysis}
\newacro{ICA}{Independent Component Analysis}
\newacro{ROI}{Region of Interest}
\newacroplural{ROI}[ROIs]{Regions of Interest}
\newacro{CT}{Computed tomography}
\newacro{DCID}{Deep Canonical Information Decomposition}
\newacro{pCCA}{Probabilistic CCA}
\newacro{MTL}{Multi-Task Learning}
\newacro{Adv. MTL}{Adversarial Multi-Task Learning}
\newacro{MTFL}{Multi-Task Feature Learning}

\newacro{CSF}{cerebrospinal fluid}
\newacro{BFM}{body fat mass}
\newacro{BMI}{body mass index}
\newacro{UKB}{UK Biobank}
\newacro{ADNI}{Alzheimer's Disease Neuroimaging Initiative}
\newacro{AD}{Alzheimer's Disease}

\newacro{CDR}{Clinical Dementia Rating}

\newacro{t.v.}{target variable}
\newacroplural{t.v}[t.v.s]{target variables}

\newacro{g.t.}{ground-truth}
\newacro{SPM}{Statistical Parametric Mapping}
\newacroplural{SPM}[SPMs]{Statistical Parametric Mappings}

\newacro{DAG}{Directed Acyclic Graph}
\newacro{SotA}{state-of-the-art}
\newacro{CMI}{Conditional Mutual Information}
\newacro{BRNet}{BRNet}
\newacro{CNC}{Correct-N-Contrast}
\newacro{ECE}{Expected Calibration Error}
\newacro{IRM}{Invariant Risk Minimization}

\title{DCID: Deep Canonical Information Decomposition}
%
%
\tocauthor{Alexander~Rakowski,Christoph~Lippert}
\toctitle{DCID: Deep Canonical Information Decomposition}
\author{Alexander~Rakowski\inst{1}[\Letter]\orcidID{0000-0002-8134-6729} \and
Christoph~Lippert\inst{1,2}\orcidID{0000-0001-6363-2556}}
\authorrunning{A. Rakowski and C. Lippert}
%
\institute{Hasso Plattner Institute for Digital Engineering, University of Potsdam, Germany \and
Hasso Plattner Institute for Digital Health at Mount Sinai, New York, United States
\email{$\{$alexander.rakowski,christoph.lippert$\}$@hpi.de}
}

\maketitle              
\begin{abstract}
We consider the problem of identifying the signal shared between two one-dimensional \textit{target} variables, in the presence of additional multivariate observations.
\ac{CCA}-based methods have traditionally been used to identify shared variables, however, they were designed for multivariate targets and only offer trivial solutions for univariate cases. 
In the context of \ac{MTL}, various models were postulated to learn features that are sparse and shared across multiple tasks.
However, these methods were typically evaluated by their predictive performance.
To the best of our knowledge, no prior studies systematically evaluated models in terms of correctly recovering the shared signal.
Here, we formalize the setting of univariate shared information retrieval, and propose ICM, an evaluation metric which can be used in the presence of ground-truth labels, quantifying 3 aspects of the learned shared features.
We further propose \ac{DCID} - a simple, yet effective approach for learning the shared variables.
We benchmark the models on a range of scenarios on synthetic data with known ground-truths and observe \ac{DCID} outperforming the baselines in a wide range of settings.
Finally, we demonstrate a real-life application of \ac{DCID} on brain \ac{MRI} data, where we are able to extract more accurate predictors of changes in brain regions and obesity.
The code for our experiments as well as the supplementary materials are available at \url{https://github.com/alexrakowski/dcid}

\keywords{Shared Variables Retrieval \and CCA \and Canonical Correlation Analysis}
\end{abstract}
\newcommand{\Real}{\mathbb{R}}
\newcommand{\dimX}{l}
\newcommand{\dimZ}{k}

\newcommand{\yRV}{Y}
\newcommand{\xRV}{\mathbf{X}}
\newcommand{\yOne}{Y_1}
\newcommand{\yTwo}{Y_2}
\newcommand{\yi}{Y_i}
\newcommand{\topC}{n}
\newcommand{\thresholdC}{T}

\newcommand{\sSharedSet}{\mathbf{S}_{share}}
\newcommand{\sIndivSet}{\mathbf{S}_{indiv}}
\newcommand{\sSharedSingle}{S}
\newcommand{\sSharedCandidate}{\hat{\mathbf{S}}_{share}}
\newcommand{\zCandidate}{\hat{\mathbf{Z}}}
\newcommand{\zCandidateVec}{\hat{\mathbf{z}}}

\newcommand{\LossFn}{\mathcal{L}}
\newcommand{\Dataset}{\mathcal{D}}
\newcommand{\xVec}{\mathbf{x}}
\newcommand{\yVec}{\mathbf{y}}
\newcommand{\zVec}{\mathbf{z}}
\newcommand{\cVec}{\mathbf{c}}
\newcommand{\transfShared}{f^{\star}}
\newcommand{\ccaWeightOne}{U}
\newcommand{\ccaWeightTwo}{V}
\newcommand{\transfRepr}{f}
\newcommand{\transfClassif}{g}
\newcommand{\transfComposed}{h}

\acresetall
\section{Introduction}
\begin{figure}
\begin{adjustwidth*}{}{-8em} 
\centering
   {
   \begin{tikzpicture}
   \tikzset{
   > = stealth,
    hidden/.style = {
        draw = black,
        shape = circle,
        minimum size = 16pt,
        inner sep = 0pt
    }
   }
   
\tikz{
    \node (y1) at (0,0) {$\yOne$};
    \node (y2) at (2,0) {$\yTwo$};
    \node[hidden] (Z) at (1,1) {$\mathbf{Z}$};
    \node[hidden] (z1) at (-1,1) {$\mathbf{Z_1}$};
    \node[hidden] (z2) at (3,1) {$\mathbf{Z_2}$};
    \node (x) at (1,2.5) {$\mathbf{X}$};
    
    \path[->] (Z) edge (y1);
    \path[->] (Z) edge (y2);
    \path[->] (z1) edge (y1);
    \path[->] (z2) edge (y2);
    \path[->] (Z) edge (x);
    \path[->] (z1) edge (x);
    \path[->] (z2) edge (x);
}
    \end{tikzpicture}  
    }
    \end{adjustwidth*}
  {\caption{
A \acf{DAG} representing the graphical model used in our setting.
Unobserved variables are denoted in circles.
The two univariate random variables $\yOne, \yTwo$ are generated by their \textbf{individual} ancestors $\mathbf{Z_1}, \mathbf{Z_2}$, and a \textbf{shared} ancestor $\mathbf{Z}$.
The multivariate $\mathbf{X}$ is generated by all 3 latent variables.
The following pairs of variables are independent under this model: $\mathbf{Z} \perp \mathbf{Z_1}$, $\mathbf{Z} \perp \mathbf{Z_2}$, $\mathbf{Z_1} \perp \mathbf{Z_2}$, $\mathbf{Z_1} \perp \yTwo$ and $\mathbf{Z_2} \perp \yOne$.
  } \label{fig:dag}}
\end{figure}
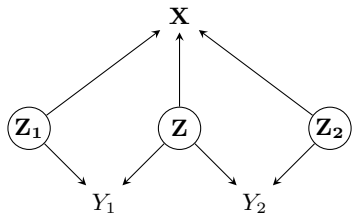

In this paper, we approach the problem of isolating the \textit{shared} signal $\mathbf{Z}$ associated with two scalar \textit{target variables} $Y_1$ and $Y_2$, from their \textit{individual} signals $\mathbf{Z_1}$ and $\mathbf{Z_2}$, by leveraging additional, high-dimensional observations $\mathbf{X}$ (see Figure~\ref{fig:dag} for the corresponding graphical model).
Analyzing the relationships between pairs of variables is ubiquitous in biomedical or healthcare studies.
However, while biological traits are often governed by complex processes and can have a wide range of causes,
we rarely have access to the fine-grained, low-level signals constituting the phenomena of interest.
Instead, we have to either develop hand-crafted quantities based on prior knowledge, which is often a costly process, or resort to high-level, ``aggregate'' measurements of the world.
If the underlying signal is weak, it might prove challenging to detect associations between such aggregated variables.
On the other hand, in many fields, we have access to high-dimensional measurements, such as medical scans or genome sequencing data, which provide rich, albeit ``unlabeled'' signal.
We propose to leverage such data to \textbf{decompose} the traits of interest into their \textbf{shared} and \textbf{individual} parts, allowing us to better quantify the relationships between them.

\ac{pCCA} was one of the early approaches to learning the shared signal between pairs of \acp{r.v.}~\cite{bach2005probabilistic,klami2008probabilistic}.
However, its effectiveness is limited to multivariate observations - 
for scalar variables we can only learn the variables themselves, up to multiplication by a constant.
A variety of methods from the field of \ac{MTFL} learn feature representations of $\mathbf{X}$, which should be sparse and shared across tasks~\cite{argyriou2006multi,kumar2012learning,yang2016deep}.
These models are typically evaluated by their predictive performance, and the shared features are rather a means of improving predictions, than a goal itself.
To the best of our knowledge, no studies exist which systematically quantify how accurate are such models in recovering the signal shared between tasks.

To this end, we define the ICM score, which evaluates 3 aspects of learned shared features: \textit{informativeness}, \textit{completeness}, and \textit{minimality}, when ground-truth labels are available.
Furthermore, we propose \ac{DCID}, an approach utilizing \ac{DNN} feature extractors and \ac{CCA} to learn the variables $\mathbf{Z}$ shared between traits.
\ac{DCID} approximates the traits of interest with \ac{DNN} classifiers and utilizes their latent features as multivariate decompositions of the traits.
It then identifies the shared factors by performing \ac{CCA} between the two sets of latent representations and retaining the most correlated components (see Figure~\ref{fig:method} for a graphical overview of the method).

Our contributions can be summarized as follows:
\begin{enumerate}
    \item We define the ICM score, which allows evaluation of learned shared features in the presence of ground-truth labels (Section~\ref{sec:problem_setting})
    \item We propose \ac{DCID}, a method leveraging \ac{DNN} classifiers and \ac{CCA} to learn the shared signal (Section~\ref{sec:method}).
    \item We benchmark the proposed model, along several baselines, on a range of scenarios with synthetic data, and analyze their performance wrt. different properties of the underlying ground-truth (Sections~\ref{ssec:benchmark_z} and~\ref{ssec:exp_variance}).
    \item Finally, we demonstrate a real-life use-case of the proposed method, by applying it on a dataset of brain \ac{MRI} data to better quantify the relationships between brain structures and obesity (Section~\ref{ssec:exp_brain_mri}).
\end{enumerate}

\section{Related Work}
\subsection{\acf{CCA}}
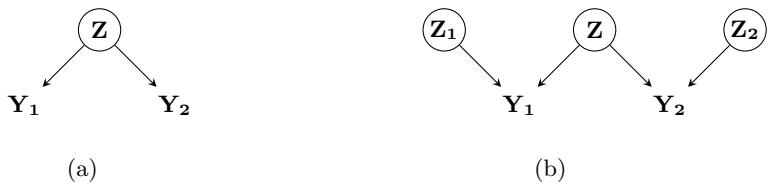
\begin{figure}
     \begin{subfigure}[b]{0.49\textwidth}
\begin{tikzpicture}
   \tikzset{
   > = stealth,
    hidden/.style = {
        draw = black,
        shape = circle,
        minimum size = 16pt,
        inner sep = 0pt
    }
   }
   
\tikz{
    \node (y1) at (0,0) {$\mathbf{Y_1}$};
    \node (y2) at (2,0) {$\mathbf{Y_2}$};
    \node[hidden] (Z) at (1,1) {$\mathbf{Z}$};
    \node (z1) at (-1,1) {};
    
    \path[->] (Z) edge (y1);
    \path[->] (Z) edge (y2);
}
    \end{tikzpicture}  
\caption{}\label{fig:dag_cca_1}
     \end{subfigure}
     \hfill
     \begin{subfigure}[b]{0.49\textwidth}
\begin{tikzpicture}
   \tikzset{
   > = stealth,
    hidden/.style = {
        draw = black,
        shape = circle,
        minimum size = 16pt,
        inner sep = 0pt
    }
   }
   
\tikz{
    \node (y1) at (0,0) {$\mathbf{Y_1}$};
    \node (y2) at (2,0) {$\mathbf{Y_2}$};
    \node[hidden] (Z) at (1,1) {$\mathbf{Z}$};
    \node[hidden] (z1) at (-1,1) {$\mathbf{Z_1}$};
    \node[hidden] (z2) at (3,1) {$\mathbf{Z_2}$};
    
    \path[->] (Z) edge (y1);
    \path[->] (Z) edge (y2);
    \path[->] (z1) edge (y1);
    \path[->] (z2) edge (y2);
}
    \end{tikzpicture}  
         \caption{}
         \label{fig:dag_cca_2}
     \end{subfigure}
\caption{
Two probabilistic interpretations of \acf{CCA}.
In ($a$), the observed variables are different, noisy views (linear transformations) of the same underlying variable $\mathbf{Z}$.
In ($b$), two additional, view-specific latent variables $\mathbf{Z_1}$ and $\mathbf{Z_2}$ are introduced, which can be interpreted as modeling the uncertainty of $p(\mathbf{Y}|\mathbf{Z})$.
} \label{fig:dag_cca}
\end{figure}
\ac{CCA} is a statistical technique operating on pairs of multivariate observations~\cite{hotelling1992relations,hardoon2004canonical}.
It is similar to \ac{PCA}~\cite{pearson1901liii}, in that it finds linear transformations of the observations, such that the resulting variables are uncorrelated.
Specifically, for a pair of observations $\mathbf{Y_1} \in \Real^{n\times p}$ and $\mathbf{Y_2} \in \Real^{n\times q}$, it finds linear transformations $U \in \Real^{p \times d},\ V \in \Real^{q \times d},\ d=min\{p,q\}$ which maximize the correlation between the consecutive pairs of the resulting variables $\mathbf{C_1} = \mathbf{Y_1} U,\ \mathbf{C_2} = \mathbf{Y_2} V$.
In the probabilistic interpretation of \ac{CCA}~\cite{bach2005probabilistic}, one can interpret the observed variables as two different views of the latent variable $\mathbf{Z}$ (Figure~\ref{fig:dag_cca_1}).
This interpretation is extended in~\cite{klami2008probabilistic} to include view-specific variables $\mathbf{Z_1}$ and $\mathbf{Z_2}$ (Figure~\ref{fig:dag_cca_2}), which is the closest to our setting.

\subsection{\acf{MTL}}
\ac{MTL} is a machine learning paradigm, where models are fitted to several tasks simultaneously, with the assumption that such joint optimization will lead to better generalization for each task~\cite{caruana1998multitask,zhang2021meat,ruder2017overview}.
In \acf{MTFL}, one aims to learn a low-dimensional representation of the input data, that is shared across tasks~\cite{argyriou2006multi,scholkopf2007multi}.
The early approaches for \ac{MTFL} worked with linear models and were based on imposing constraints on the matrix of model parameters, such as sparsity or low-rank factorization~\cite{argyriou2006multi,argyriou2008convex,kumar2012learning}.
Modern approaches extend \ac{MTFL} to \ac{DNN} models by using tensor factorization in place of matrix factorization~\cite{wimalawarne2014multitask,yang2016deep} or by employing adversarial training to learn task-invariant, and task-specific features~\cite{bousmalis2016domain,shinohara2016adversarial,liu2017adversarial}.

\section{Univariate Shared Information Retrieval}\label{sec:problem_setting}
In this section, we formalize the problem setting (Section~\ref{ssec:problem_setting}) and define 3 quantities measuring different aspects of the learned shared representations, which constitute the model evaluation procedure (Section~\ref{ssec:evaluation}).

\subsection{Problem Setting}\label{ssec:problem_setting}
In our setting, we observe two univariate \acp{r.v.} $\yOne, \yTwo \in \Real$, which we will refer to as the \textit{target variables}, and a multivariate \ac{r.v.} $\mathbf{X} \in \Real^{\dimX}$.
We further define 3 unobserved, multivariate, and pairwise-independent \textit{latent variables} $\mathbf{Z, Z_1, Z_2} \in \Real^{\dimZ}$, which generate the observed variables.
We will refer to $\mathbf{Z_1}$ and $\mathbf{Z_2}$ as the \textit{individual variables}, and to $\mathbf{Z}$ as the \textit{shared variables}.
The main assumption of the model is that the individual variables $\mathbf{Z_1}$ and $\mathbf{Z_2}$ are each independent from one of the target variables, i.e., $\mathbf{Z_1} \perp \yTwo$ and $\mathbf{Z_2} \perp \yOne$, while the shared variable $\mathbf{Z}$ is generating all the observed \acp{r.v.}, i.e., $Y_1, Y_2$ and $\mathbf{X}$.
The corresponding graphical model is shown in Figure~\ref{fig:dag}.
Similar to the \ac{pCCA} setting, we assume additivity of the effects of the shared and individual variables on $Y_i$, i.e.:
\begin{equation}\label{eq:target_decomposition}
    Y_i = \psi_i(\mathbf{Z}) + \phi_i(\mathbf{Z_i}),\ i \in \{1,2\}
\end{equation}
where $\psi_i$ and $\phi_i$ are arbitrary functions $\Real^{\dimZ} \mapsto \Real$.

Our task of interest is then predicting the shared variable $\mathbf{Z}$ given the observed \acp{r.v.}, i.e., learning an accurate model of $p(\mathbf{Z}| \yOne, \yTwo, X)$, \textbf{without access to $\mathbf{Z, Z_1, Z_2}$ during training}.

\subsection{Evaluating the Shared Representations}\label{ssec:evaluation}
\newcommand{\LossName}{ICM}
\newcommand{\LossInform}{\mathcal{L}_{info}}
\newcommand{\LossComp}{\mathcal{L}_{comp}}
\newcommand{\LossMinim}{\mathcal{L}_{min}}
While in practical scenarios we assume that the latent variables remain unobserved, to benchmark how well do different algorithms recover $\mathbf{Z}$, we need to test them in a controlled setting, where all ground-truth variables are available at least during test time.
Let $\Dataset=\{\mathbf{x}^{(i)}, y_1^{(i)}, y_2^{(i)}, \mathbf{z}^{(i)}, \mathbf{z}_1^{(i)}, \mathbf{z}_2^{(i)}\}^N_{i=1}$ be a ground-truth dataset, and $\hat{\mathbf{z}}=\{ \hat{\mathbf{z}}^{(i)}\}^N_{i=1}$ be the learned shared representations.
We will denote by $[\mathbf{x}, \mathbf{y} ]$ the column-wise concatenation of $\mathbf{x}$ and $\mathbf{y}$, and by $R^2(\mathbf{x}, \mathbf{y})$ the ratio of variance explained (i.e., the coefficient of determination) by fitting a linear regression model of $\mathbf{x}$ to $\mathbf{y}$:
\begin{equation}
R^2(\mathbf{x}, \mathbf{y}) = 1 - \frac{1}{d} \sum^d_{j=1} \frac{\sum_i (e_j^{(i)})^2}{\sum_i (y_j^{(i)} - \bar{y}_j)^2} 
\end{equation}
where $e^{(i)}_j = \hat{y_j}^{(i)} - y^{(i)}_j$ are residuals of the model of the $j$-th dimension of $\mathbf{y}$.

Inspired by the DCI score~\cite{eastwood2018framework} from the field of disentangled representation learning, we define the following requirements for a learned representation $\zCandidate$ as correctly identifying the shared variable $\mathbf{Z}$:
\begin{enumerate}
    \item \textbf{Informativeness}: $\mathbf{Z}$ should be predictable from $\zCandidate$. We measure this as the ratio of variance explained by a model fitted to predict $\mathbf{Z}$ from $\zCandidate$:
    \begin{equation}
        \LossInform(\hat{\mathbf{z}}, \mathbf{z}) = R^2(\hat{\mathbf{z}}, \mathbf{z})
    \end{equation}
    \item \textbf{Compactness}: $\mathbf{Z}$ should be sufficient to predict $\zCandidate$. We measure this as the ratio of variance explained by a model fitted to predict $\zCandidate$ from $\mathbf{Z}$:
    \begin{equation}
        \LossComp(\mathbf{z}, \hat{\mathbf{z}}) = R^2(\mathbf{z}, \hat{\mathbf{z}})
    \end{equation}
    \item \textbf{Minimality}: $\zCandidate$ should only contain information about $\mathbf{Z}$. We measure this as one minus the ratio of variance explained by a model fitted to predict the individual variables $\mathbf{Z_1}$ and $\mathbf{Z_2}$ from $\zCandidate$:
    \begin{equation}
        \LossMinim(\hat{\mathbf{z}}, \mathbf{z_1}, \mathbf{z_2}) = 1 - R^2(\hat{\mathbf{z}}, [\mathbf{z_1}, \mathbf{z_2}])
    \end{equation}
\end{enumerate}
The final score, $\LossName$, is given as the product of the individual scores:
\begin{equation}
\begin{split}
    \LossName(\hat{\mathbf{z}}, \Dataset) = \LossInform(\hat{\mathbf{z}}, \mathbf{z}) \cdot \LossComp(\mathbf{z}, \hat{\mathbf{z}}) \cdot \LossMinim(\hat{\mathbf{z}}, \mathbf{z_1}, \mathbf{z_2})
\end{split}
\end{equation}
and takes values in $[0, 1]$, with 1 being a perfect score, identifying $\mathbf{Z}$ up to a rotation.

Note that minimality might seem redundant given compactness - if $\mathbf{Z}$ explains all the variance in $\zCandidate$, then, since $\mathbf{Z} \perp \mathbf{Z_1}, \mathbf{Z_2}$, $\zCandidate$ would contain no information about $\mathbf{Z_1}$ or $\mathbf{Z_2}$.
However, if we accidentally choose the dimensionality of $\zCandidate$ to be much higher than that of $\mathbf{Z}$, a model can ``hide'' information about $\mathbf{Z_1}$ and $\mathbf{Z_2}$ by replicating the information about $\mathbf{Z}$ multiple times, e.g., $\zCandidate = \{ \mathbf{Z_1}, \mathbf{Z_2}, \mathbf{Z}, \ldots \mathbf{Z} \}$.
This would result in a perfect informativeness and an almost perfect compactness score, but a low minimality score.

\section{Method: Deep Canonical Information Decomposition}\label{sec:method}
In this section, we outline the difficulty in tackling the problem formulated above with \ac{CCA} (Section~\ref{ssec:cca_shortcomings}), and describe an algorithm for solving it by exploiting the additional observed variable $\mathbf{X}$ (Section~\ref{ssec:dcid}).

\subsection{Limitations of the \ac{CCA} Setting}\label{ssec:cca_shortcomings}
Without $\mathbf{X}$, our setting can be seen as a special case of \acf{pCCA}~\cite{klami2008probabilistic}, where the observed variables (``views'' of the data) have a dimensionality of one (see Figure~\ref{fig:dag_cca_2}).
If we assume non-empty $\mathbf{Z}$, then, by the \ac{pCCA} model, $\mathbf{Z_1}$ and $\mathbf{Z_2}$ would have a dimensionality of zero, resulting in degenerate solutions in form of:
\begin{equation}
\begin{split}
    \hat{Z} = \alpha Y_1 \\ \lor \\ \hat{Z} = \alpha Y_2 \\
  \alpha \neq 0
\end{split}
\end{equation}
as the only linear transformations of univariate $Y_1$ and $Y_2$ are the variables themselves, up to scalar multiplication.

\subsection{\acf{DCID}}\label{ssec:dcid}
\begin{figure}
\includegraphics[width=\textwidth]{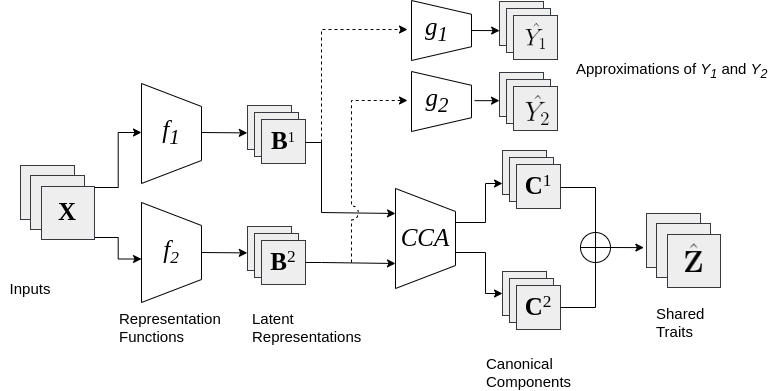}
\caption{
A visual illustration of the \acf{DCID} model. 
The target variables $Y_1$ and $Y_2$ are approximated by fitting \ac{DNN} predictors on the high-dimensional data $\mathbf{X}$.
Outputs of the penultimate layers of the networks are then used as multivariate decompositions of $Y_1$ and $Y_2$, and fed into \ac{CCA} to estimate the shared signal $\zCandidate$.
} \label{fig:method}
\end{figure}

In order to find $\mathbf{Z}$, we need ``unaggregated'', multivariate views of $Y_1$ and $Y_2$.
To achieve this, we leverage the high-dimensional observations $\mathbf{X}$, e.g., images, to learn decompositions of $Y_1, Y_2$ as transformations of $\mathbf{X}$.
Specifically, we assume that both $Y_i$ can be approximated as transformations of $\mathbf{X}$ with functions $h_i(\mathbf{X}) = \hat{Y_i}$.
We further assume they can be decomposed as $h_i = g_i \circ f_i$, where $f_i: \Real^{\dimX} \mapsto \Real ^{\dimZ}$, called the \textit{representation function}, can be an arbitrary, potentially nonlinear mapping, and $g_i: \Real^{\dimZ} \mapsto \Real$, called the \textit{classifier function}, is a linear combination.
The $\dimZ$-dimensional outputs of $f_i(\mathbf{X})=\mathbf{B_i}$ constitute the multivariate decompositions of $Y_i$.
Since these are no longer univariate, we can now apply the standard \ac{CCA} algorithm on $\mathbf{B_1, B_2}$ to obtain pairs of canonical variables $\mathbf{C_1, C_2} \in \Real^{\dimZ}$, sorted by the strength of their pairwise correlations, i.e., 
\begin{equation}
\forall i, j \in |\dimZ|: i < j \Rightarrow corr(\mathbf{C}_{1, i}, \mathbf{C}_{2, i}) \leq corr(\mathbf{C}_{1, j}, \mathbf{C}_{2, j})    
\end{equation}
In order to extract the most informative features, we can select the $\topC$ pairs of canonical variables with correlations above a certain threshold $\thresholdC$:
\begin{equation}\label{eq:threshold_C}
    \topC = \argmax_{i \in |\dimZ|} corr(\mathbf{C}_{1, i}, \mathbf{C}_{2, i}) > T
\end{equation}
We then take the $\zCandidate = \frac{1}{2}(\mathbf{C}_{1, 1:\topC} + \mathbf{C}_{2, 1:\topC})$ as our estimate of the shared $\mathbf{Z}$.
The complete process is illustrated in Figure~\ref{fig:method} and described step-wise in Algorithm~\ref{alg:method}.
\subsubsection{Modeling the $h_i$}
In practice, we approximate each $h_i$ by training \ac{DNN} models to minimize $\E [Y_i - h_i(\mathbf{X})]^2$, i.e., a standard \ac{MSE} objective.
\acp{DNN} are a fitting choice for modeling $h_i$, since various popular architectures, e.g., ResNet~\cite{he2016deep}, can naturally be decomposed into a nonlinear \textit{feature extractor} (our $f_i$) and a linear \textit{prediction head} (our $g_i$).

\RestyleAlgo{ruled}
\SetKwComment{Comment}{/* }{ */}
\begin{algorithm}
\caption{\ac{DCID}: Computing the shared features $\zCandidateVec$ and the prediction function $\transfShared$}
\label{alg:method}

\KwIn{$\Dataset=\{\mathbf{x}^{(i)}, y_1^{(i)}, y_2^{(i)}\}^N_{i=1}$  \Comment*[r]{Training dataset}}
\KwIn{$\LossFn(\cdot, \cdot)$  \Comment*[r]{Loss function to optimize the \acp{DNN}, e.g., $L_2$}}
\KwIn{$\thresholdC$ \Comment*[r]{Canonical correlation threshold}}
\KwOut{$\zCandidate$ \Comment*[r]{Features shared between $\yOne$ and $\yTwo$}} 
\KwOut{$\transfShared$ \Comment*[r]{Function to predict $\zCandidate$ from new data}}

$\transfRepr_1, \transfClassif_1 \leftarrow \argmin_{\transfComposed = \transfClassif \circ \transfRepr} \LossFn(\yVec_1, \transfComposed(\xVec))$ \Comment*[r]{Fit a \ac{DNN} to predict $\yVec_1$ from $\xVec$}
$\transfRepr_2, \transfClassif_2 \leftarrow \argmin_{\transfComposed = \transfClassif \circ \transfRepr} \LossFn(\yVec_2, \transfComposed(\xVec))$ \Comment*[r]{Fit a \ac{DNN} to predict $\yVec_2$ from $\xVec$}
$\mathbf{b}_1 \leftarrow \transfRepr_1(\xVec)$ \;
$\mathbf{b}_2 \leftarrow \transfRepr_2(\xVec)$ \;
$\ccaWeightOne, \ccaWeightTwo \leftarrow $ CCA$(\mathbf{b}_1, \mathbf{b}_2)$ \Comment*[r]{Compute the CCA projection matrices $\ccaWeightOne, \ccaWeightTwo$}

$\zCandidateVec \leftarrow \emptyset$ \;
$\topC \leftarrow 1$ \;
\While{
$corr(\ccaWeightOne_{\topC}^{\top} \mathbf{b}_1, \ccaWeightTwo_{\topC}^{\top} \mathbf{b}_2) > \thresholdC$
}{
$\mathbf{b}^{\topC} \leftarrow \frac{1}{2}(\ccaWeightOne_{\topC}^{\top} \mathbf{b}_1 + \ccaWeightTwo_{\topC}^{\top} \mathbf{b}_2)$ \;
$\zCandidateVec \leftarrow \zCandidateVec \cup \{\mathbf{b}^{\topC} \}$ \Comment*[r]{Add a new shared component $\sSharedSingle^{\topC}$}
$\topC \leftarrow \topC + 1$ \;
}
$\topC \leftarrow \topC - 1$ \;
$\transfShared(\cdot) \leftarrow \frac{1}{2} [\ccaWeightOne_{1:\topC}^{\top}\transfRepr_1(\cdot) + \ccaWeightTwo_{1:\topC}^{\top}\transfRepr_2(\cdot)]$ \Comment*[r]{Save the function $\transfShared$}
\end{algorithm}

\newcommand{\wCutoff}{T_{MTL}}
\section{Experiments}\label{sec:experiments}
In Section~\ref{ssec:exp_baselines} we describe the baseline models we compare against, and in Section~\ref{ssec:exp_settings} we describe the settings of the conducted experiments, such as model hyperparameters or datasets used.
In Sections~\ref{ssec:benchmark_z} and~\ref{ssec:exp_variance} we conduct experiments on synthetic data with know ground-truth - in Section~\ref{ssec:benchmark_z} we benchmark the models in terms of retrieving the shared variables $\mathbf{Z}$, and in Sections~\ref{ssec:exp_variance} we evaluate how the performance of the models degrades when the variance explained by the shared variables changes.
Finally, in Section~\ref{ssec:exp_brain_mri} we demonstrate a real-life use case of the proposed \ac{DCID} method on brain \ac{MRI} data, where the underlying ground-truth is not known.

\subsection{Baselines}\label{ssec:exp_baselines}
\paragraph{\acf{MTL}} We train a \ac{DNN} model in a standard multi-task setting, i.e., to predict both $Y_1$ and $Y_2$ with a shared feature extractor $f : \Real^{\dimX} \mapsto \Real^{\dimZ}$ and task-specific linear heads $g_1, g_2 : \Real^{\dimZ} \mapsto \Real$.
 We then select as $\zCandidate$ the set of features of $f$, for which the magnitude of normalized weights of the task-specific heads exceeds a certain threshold $\wCutoff$ for both heads, i.e.:
 \begin{equation}
     f(\cdot)_i \in \zCandidate \implies \wCutoff \leq \frac{|G_{1, i}|}{max|G_{1, :}|}\ \wedge \wCutoff \leq \frac{|G_{2, i}|}{max|G_{2, :}|}\
 \end{equation}
where $G_1, G_2$ are weight vectors of the linear heads $g_1, g_2$.
\paragraph{\acf{MTFL}} We train a multi-task \ac{DNN} as in the \ac{MTL} setting, and apply the algorithm for \textit{sparse common feature learning} of~\cite{argyriou2008convex} on the features of $f$.
This results in new sparse features $f'$ and their corresponding new prediction heads $g_1', g_2'$.
As in the above setting, we select features of $f'$ with the magnitude of normalized weights for $g_i'$ above a threshold $\wCutoff$.
\paragraph{\acf{Adv. MTL}} Introduced in~\cite{liu2017adversarial}, this model learns 3 disjoint feature spaces: 2 task-specific, \textit{private} feature spaces, and a \textit{shared} space, with features common for both tasks.
The model is trained in an adversarial manner, with the discriminator trying to predict the task from the shared features.
Additionally, it imposes an orthogonality constraint on the shared and individual spaces, forcing them to contain different information.

\subsection{Experimental Settings}\label{ssec:exp_settings}
\subsubsection{Synthetic Data}
For experiments with known ground-truth, we employed the Shapes3D dataset~\cite{3dshapes18}, which contains synthetic $64\times64$ pixel RGB images of simple 3-dimensional objects against a background, generated from 6 independent latent factors: floor hue, wall hue, object hue, scale, shape and orientation of the object, resulting in $480,000$ samples total.
We take the images as $\mathbf{X}$, and select different factors as the unobserved variables $\mathbf{Z, Z_1, Z_2}$.
As the 6 factors are the only sources of variation in the observed data $\mathbf{X}$, it allows for an accurate evaluation of model performance in terms of retrieving $\mathbf{Z}$.

We employed the encoder architecture from~\cite{locatello2019challenging} as the \ac{DNN} backbone used to learn $f_1$ and $f_2$.
The models were trained for a single pass over the dataset, with a mini-batch of size 128 using the Adam optimizer~\cite{kingma2014adam} with a learning rate of $10^{-4}$.
We repeated each experimental setting over 3 random seeds, each time splitting the dataset into different train/validation/test split with ratios of (70/15/15)\%.
For the hyperparameter sweep performed in Section~\ref{ssec:benchmark_z} we considered: 10 values evenly spread on $[0,1]$ for $\thresholdC$ of \ac{DCID}, 10 values evenly spaced on $[0,1]$ for $\wCutoff$ of \ac{MTL}, \ac{MTFL}, and \ac{Adv. MTL}, 10 values evenly spread on a logarithmic scale of $[10^{-4}, 10]$ for the $\gamma$ parameter of \ac{MTFL}, and $\gamma, \lambda \in \{0.01, 0.05, 0.1, 0.5, 1 \}$ and a learning rate of the discriminator in $\{10^{-4}, 10^{-3}\}$ for \ac{Adv. MTL}.

\subsubsection{Brain \ac{MRI} Data}\label{sssec:exp_settings_mri}
For the experiments on brain \ac{MRI} scans (Section~\ref{ssec:exp_brain_mri}) we employed data from the \acf{UKB} medical database~\cite{sudlow2015uk}.
Specifically, we selected data for participants who underwent the brain scanning procedure, self-identified as ``white-British'', and have a similar genetic ancestry, which resulted in $34,314$ data points.
As the input data $\mathbf{X}$ we took the T1-weighted structural scans, which were non-linearly registered to an MNI template~\cite{miller2016multimodal,alfaro2018image}, and downsampled them to a size of $96\times96\times96$ voxels.
For $Y_2$, we selected body mass-related measurements available in the dataset, such as the total \ac{BFM}, weight, or \ac{BMI}.
For $Y_1$, we computed the total volumes of several brain \acp{ROI}, e.g., the total volume of hippocampi or lateral ventricles, using the Synthseg software~\cite{billot_synthseg_2021}.

We employed a 3D MobileNetV2~\cite{kopuklu2019resource}, with a width parameter of 2, as the \ac{DNN} architecture used to learn $f_1$ and $f_2$.
The models were trained for 40 epochs with a mini-batch size of 12 using the Adam optimizer with a learning rate of $10^{-4}$.
For each possible pair of $Y_1$ and $Y_2$ we repeated the experiments across 3 random seeds, each time selecting a different 30\% of the samples as the test set.

\subsection{Learning the Shared Features $\mathbf{Z}$}\label{ssec:benchmark_z}
To evaluate how accurately do different models learn the underlying shared features $\mathbf{Z}$, we trained them in controlled settings with known ground-truth.
We created the latent variables from the 6 factors of the Shapes3D dataset, by randomly selecting two individual factors $Z_1, Z_2$ and one shared factor $Z$, and constructed the target variables as $Y_1 = Z_1 + Z$ and $Y_2 = Z_2 + Z$.
This resulted in 60 possible scenarios with different underlying latent variables.
To ensure a fair comparison, for each model we performed a grid search over all hyperparameters on 30 random scenarios, and evaluated it on the remaining 30 scenarios using the best found hyperparameter setting.

The resulting $\LossName$ scores are shown in Table~\ref{tab:benchmark_z}.
The proposed \ac{DCID} model performed best both in terms of the final $\LossName$ score, as well as the individual scores.
The \ac{MTL} and \ac{MTFL} methods performed similarly in terms of the $\LossName$ score, with \ac{MTL} achieving higher \textit{informativeness}, and \ac{MTFL} a lower \textit{minimality} score.
The \ac{Adv. MTL} model had the lowest $\LossName$, performing well only in terms of \textit{minimality}.

\begin{table}
\caption{
$\LossName$ scores of models trained on the Shapes3D dataset to retrieve the shared $\mathbf{Z}$.
Reported are the mean and standard deviation of each score over 90 runs per model (30 scenarios $\times$ 3 random seeds).
}\label{tab:benchmark_z}
\begin{tabular}{|l|c|c|c|c|}
\hline
       \textbf{Model} &     \textbf{ICM $\uparrow$} & \textbf{Informativeness $\uparrow$} & \textbf{Compactness $\uparrow$} & \textbf{Minimality $\downarrow$} \\
\hline
    \textbf{Adv. MTL} &          $0.06\ (\pm 0.05)$ &                  $0.22\ (\pm 0.11)$ &              $0.22\ (\pm 0.11)$ &               $0.03\ (\pm 0.02)$ \\
         \textbf{MTL} &          $0.18\ (\pm 0.14)$ &                  $0.65\ (\pm 0.19)$ &              $0.33\ (\pm 0.19)$ &               $0.23\ (\pm 0.12)$ \\
        \textbf{MTFL} &          $0.19\ (\pm 0.15)$ &                  $0.47\ (\pm 0.23)$ &              $0.37\ (\pm 0.22)$ &               $0.12\ (\pm 0.08)$ \\
\textbf{DCID} (ours)  & $\mathbf{0.62}\ (\pm 0.15)$ &         $\mathbf{0.85}\ (\pm 0.07)$ &     $\mathbf{0.73}\ (\pm 0.17)$ &      $\mathbf{0.01}\ (\pm 0.02)$ \\
\hline
\end{tabular}
\end{table}

\subsection{Variance Explained by $\mathbf{Z}$ and Model Performance}\label{ssec:exp_variance}
\newcommand{\varRatioDecrease}{\tau}
\newcommand{\varRatioDiffering}{\kappa}
We further investigated how the amount of variance in $Y_1, Y_2$ explained by $\mathbf{Z}$ influences model performance, with two series of experiments.
In the first one, we controlled $\varRatioDecrease$, the ratio of variance in $Y_1, Y_2$ explained by the shared variables $\mathbf{Z}$ to the variance explained by the individual variables $\mathbf{Z_1, Z_2}$, i.e.:
\begin{equation}
    \varRatioDecrease = \frac{R^2(\mathbf{Z}, [Y_1, Y_2 ])}{R^2([\mathbf{Z_1, Z_2}], [Y_1, Y_2])}
\end{equation}
We created 15 different base scenarios, each time selecting a different pair of variables as $\mathbf{Z_1, Z_2}$, and the remaining 4 as $\mathbf{Z}$.
For each scenario we then varied $\varRatioDecrease$ 17 times on a logarithmic scale from $0.1$ to $10$, and trained models using their best hyperparameter settings from Section~\ref{ssec:benchmark_z}.

\begin{figure}
\includegraphics[width=\textwidth]{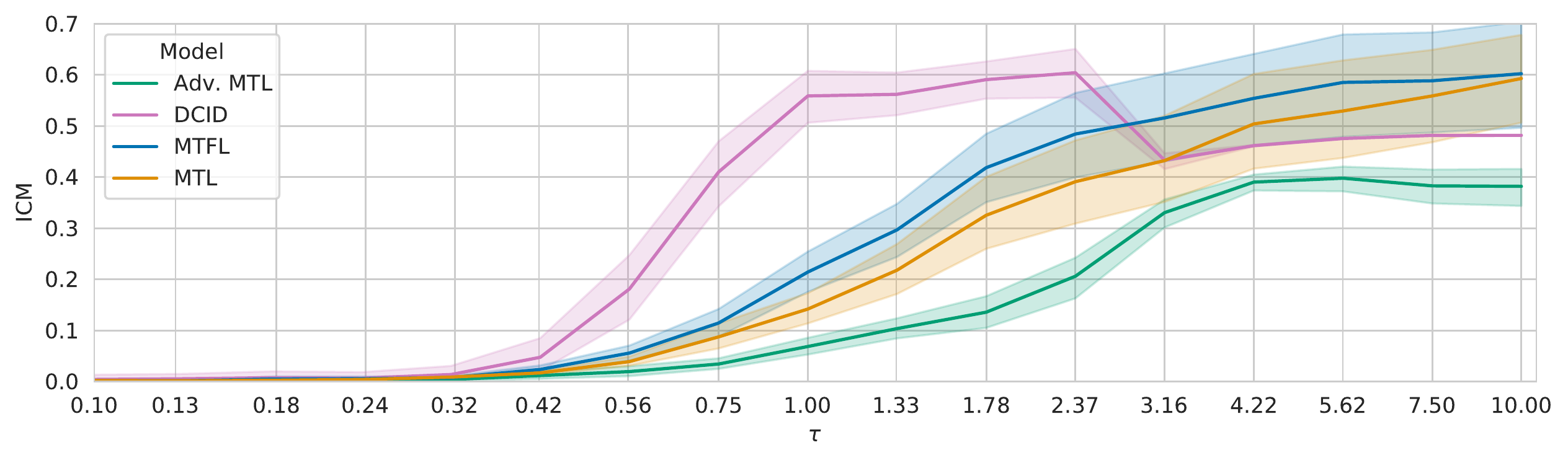}
\caption{
\textit{ICM} scores (y-axis) of different models plotted against $\varRatioDecrease$, the ratio of variance in $Y_1, Y_2$ explained by $\mathbf{Z}$ to the variance explained by $\mathbf{Z_1, Z_2}$ (x-axis, logarithmic scale).
} \label{fig:exp_decreasing_var}
\end{figure}

We plot the resulting \textit{ICM} scores against $\varRatioDecrease$ in Figure~\ref{fig:exp_decreasing_var}.
Firstly, all the models fail to recover $\mathbf{Z}$ for $\varRatioDecrease \leq 0.3$, i.e., when the signal of $\mathbf{Z}$ is weak.
For $\varRatioDecrease \in [0.3, 2.3]$ the \ac{DCID} model outperforms the baselines by a wide margin, even for $\varRatioDecrease < 1$.
The MTL models begin to recover $\mathbf{Z}$ only when it dominates the signal in the target variables.
Interestingly, the performance of \ac{DCID} drops suddenly for $\varRatioDecrease = 2.37$, and is outperformed by the \ac{MTL} and \ac{MTFL} models for $\varRatioDecrease > 2.8$.
This is a surprising behavior, and we observed it occur independently of values of the threshold $\thresholdC$ (see Figure 1 of the supplementary material).

In the second scenario, we controlled $\varRatioDiffering$, the ratio of variance explained by $\mathbf{Z}$ in $Y_1$ to the variance explained in $Y_2$, i.e.:
\begin{equation}
    \varRatioDiffering = \frac{R^2(\mathbf{Z}, Y_1)}{R^2(\mathbf{Z}, Y_2)}
\end{equation}
We created 60 base scenarios, similarly as in Section~\ref{ssec:benchmark_z}, and for each we varied $\varRatioDiffering$ 6 times evenly on the scale from $0.1$ to $1$.
Again, we selected the model hyperparameters that performed best in Section~\ref{ssec:benchmark_z}.

\begin{figure}
\includegraphics[width=\textwidth]{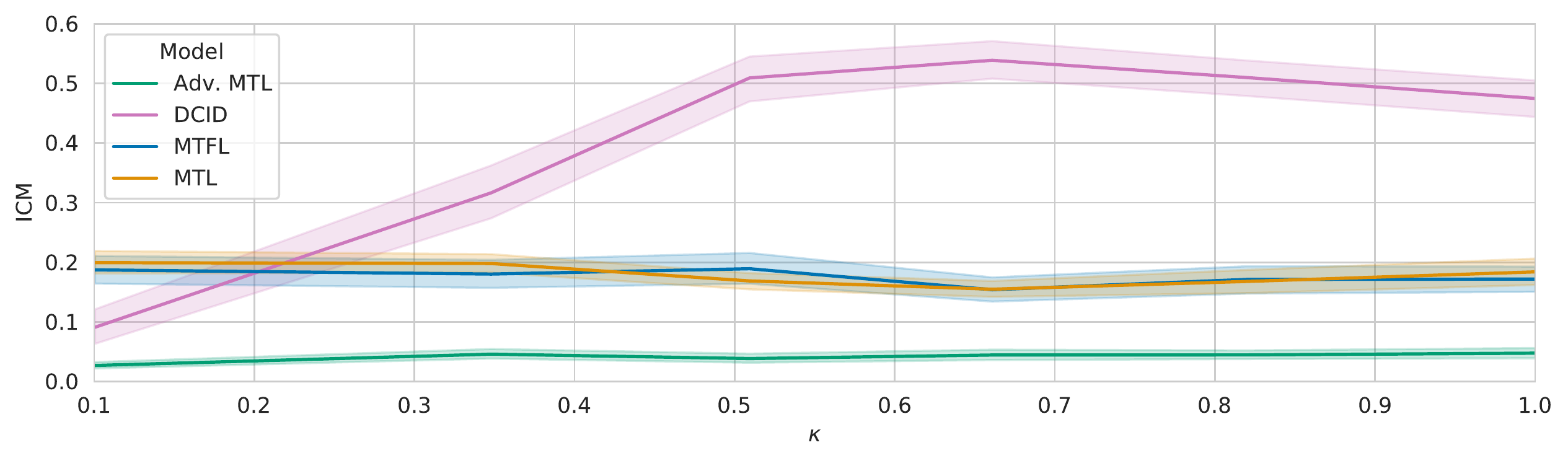}
\caption{
\textit{ICM} scores (y-axis) of different models plotted against $\varRatioDiffering$, the ratio of variance explained by $\mathbf{Z}$ in $Y_1$ to the variance explained in $Y_2$ (x-axis).
For $\varRatioDiffering=1$ the shared variables explain the same amount of variance in both target variables.
} \label{fig:exp_differing_var}
\end{figure}

We plot the resulting \textit{ICM} scores against $\varRatioDiffering$ in Figure~\ref{fig:exp_differing_var}.
For $\varRatioDiffering \in [0.5, 1]$ the \ac{DCID} model retains a consistent performance.
For values of $\varRatioDiffering$ below $0.5$ its performance decreases linearly, for $\varRatioDiffering \leq 0.2$ achieving lower \textit{ICM} scores than the \ac{MTL} and \ac{MTFL} models.
The scores in the low $\varRatioDiffering$ regime are higher, however, than the scores for low $\varRatioDecrease$ values, indicating that while \ac{DCID} performs best when $\mathbf{Z}$ explains a large amount of variance in both target variables, it is also beneficial if at least one of the target variables is strongly associated with $\mathbf{Z}$.
The baseline models, while achieving lower scores overall, seem to have their performance hardly affected by changes in $\varRatioDiffering$.

\subsection{Obesity and the Volume of Brain \acfp{ROI}}\label{ssec:exp_brain_mri}
\subsubsection{Background} The occurrence of neuropsychiatric disorders is associated with a multitude of factors.
For example, the risk of developing dementia can depend on age~\cite{chen2009risk,stephan2018subjective}, ethnicity~\cite{chen2018racial,shiekh2021ethnic}, 
 or genetic~\cite{strittmatter1993apolipoprotein,emrani2020apoe4}, vascular~\cite{cherbuin2015blood,prabhakaran2019blood}, and even dietary~\cite{zhang2021meat,frausto2021dietary} factors.
However, only a subset of these factors are modifiable.
Being aware of the genetic predisposition of an individual for developing a disorder does not directly translate into possible preventive actions.
On the other hand, mid-life obesity is a known factor for dementia, which can be potentially acted upon~\cite{gorospe2007risk,baumgart2015summary}.
Several studies analyzed the statistical relations between brain \acp{ROI} and obesity~\cite{raji2010brain,driscoll2012midlife,dekkers2019obesity}.
A natural limitation of these studies is the fact that they work with ``aggregated'' variables, quantifying obesity or \ac{ROI} volumes as single values, potentially losing information about the complex traits.
Phenomena such as the ``obesity paradox'', where obesity can have both adverse and positive effects~\cite{monda2017obesity}, indicate the need for deeper dissecting the variables of interest and the connections between them.

\subsubsection{Analysis using \ac{DCID}} We approach this problem by estimating the shared signal $\mathbf{Z}$ between $Y_1$, the \acf{BFM}, and $Y_2$, the volume of different brain \acp{ROI}.
We trained several \ac{DCID} models on the \ac{UKB} data to predict \ac{BFM} and volumes of the following \acp{ROI}: brain stem, \ac{CSF}, subcortical gray matter, ventricles, and the hippocampus.
Additionally, we trained models for $Y_1$ being the body weight, or \ac{BMI}, and report results for these in Section 2 of the supplementary material.
Since the main interest lies in the effect of obesity on the \acp{ROI}, we constructed ``surrogate'' variables of $Y_1$, denoted by $\psi_1(\mathbf{Z})$ (see Equation~\ref{eq:target_decomposition}), which isolate the shared signal in $Y_1$ from the individual one.
This is a conservative approach since it only utilizes features of the model trained to predict $Y_1$, with the information about $Y_2$ used only to rotate the features and extract the shared dimensions.

\begin{figure}
     \begin{subfigure}[b]{0.49\textwidth}
         \centering
         \includegraphics[width=\textwidth]{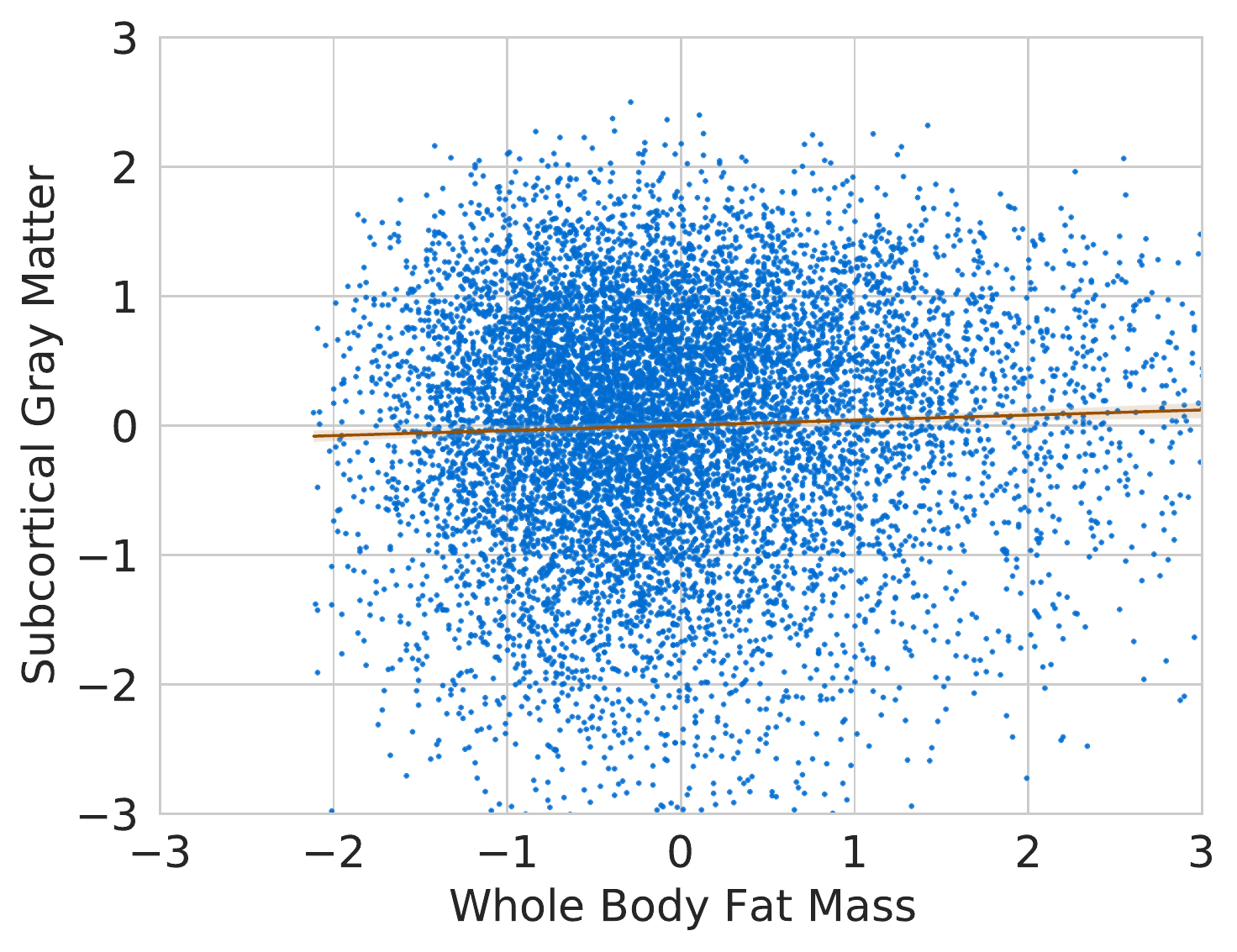}
         \caption{}
     \end{subfigure}
     \hfill
     \begin{subfigure}[b]{0.49\textwidth}
         \centering
                  \includegraphics[width=\textwidth]{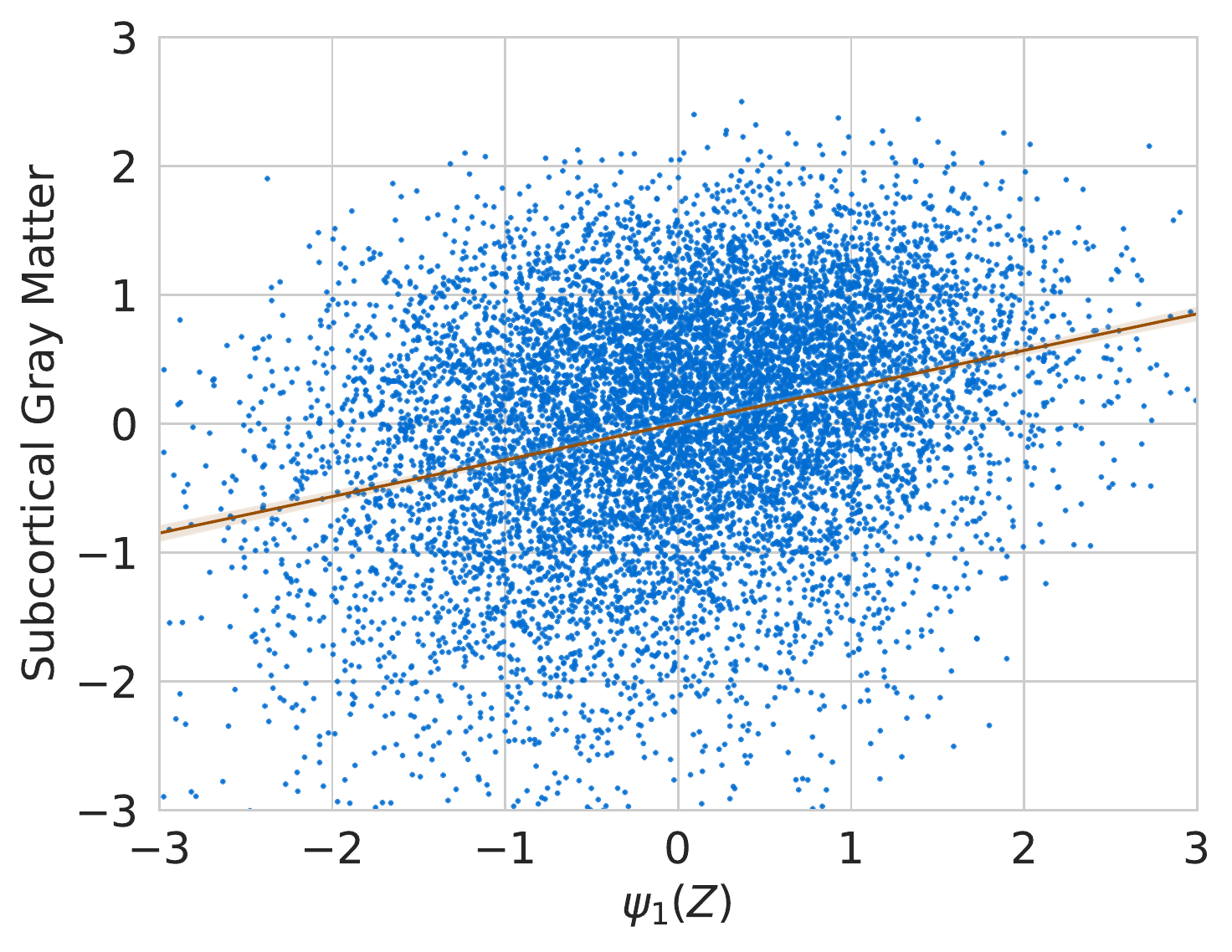}
         \caption{}
     \end{subfigure}
\caption{
Volumes of subcortical gray matter plotted against body fat mass ($a$) and against the surrogate variable $\psi_1(\mathbf{Z})$ ($b$), for a single trained model.
All variables were standardized to a z-score before plotting.
} \label{fig:brain_wbfm_surrogate}
\end{figure}

First, we demonstrate how $\psi_1(\mathbf{Z})$ allows for more accurate estimates of change in the \acp{ROI}, since it ignores the signal in $Y_1$ which is independent of $Y_2$.
We fitted $\mathbf{Z}$ on the training data by selecting shared components with a threshold $\thresholdC > 0.2$.
We then obtained predictions of $\psi_1(\mathbf{Z})$ on the test set and computed their correlation with the \ac{ROI}.
We report the results for all the \acp{ROI} in Table~\ref{tab:brain_wbfm_surrogate}, and plot \ac{BFM} and $\psi_1(\mathbf{Z})$ against the volume of the subcortical gray matter for a single model in Figure~\ref{fig:brain_wbfm_surrogate}.
For all \acp{ROI} the surrogate variable is correlated stronger than \ac{BFM}, up to 8-fold for the ventricles, while retaining the sign of the coefficient.
The smallest gains seem to be achieved for \ac{CSF}, where the spread of coefficients over different runs is also the highest.

\begin{figure}
\includegraphics[width=\textwidth]{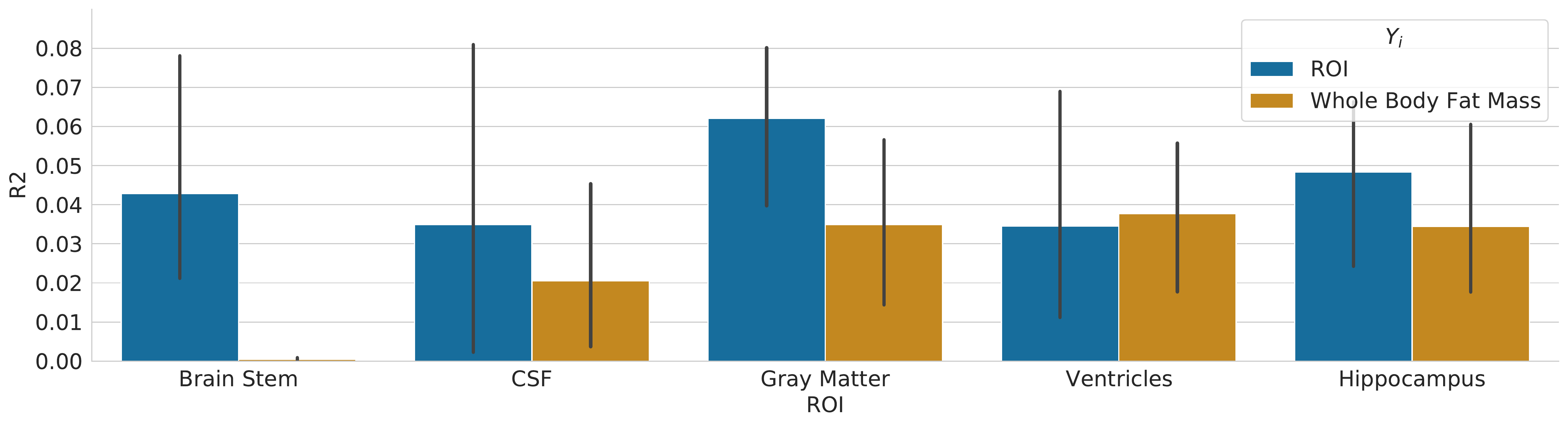}
\caption{
Ratio of total variance explained by the surrogate variable $\psi_1(\mathbf{Z})$ in different brain \acfp{ROI} (blue bars) and in \acf{BFM} (yellow bars).
} \label{fig:brain_wbfm_r2}
\end{figure}

Secondly, we show how obtaining $\psi_1(\mathbf{Z})$ allows us to estimate the variance explained separately in $Y_1$ and $Y_2$, which is not possible by merely computing the correlation coefficient between $Y_1$ and $Y_2$.
We plot the ratio of explained variance for each \ac{ROI} in Figure~\ref{fig:brain_wbfm_r2}.
While $\psi_1(\mathbf{Z})$ explains a similar amount of variance for ventricles and \ac{BFM}, we can see bigger disparities for other \acp{ROI}, especially for the brain stem, where the variance explained in \ac{BFM} is negligible.
This might indicate that predictions of the the brain stem volume from \ac{BFM} would be less reliable than predictions of other \acp{ROI}.

\begin{table}
\caption{
Pearson correlation coefficients between volumes of \acfp{ROI} in brain \ac{MRI} scans (columns) and two variables - $Y_1$, being the measurements of total body fat mass (first row), and a surrogate variable $\psi_1(\mathbf{Z})$, isolating the signal of the shared variables $\mathbf{Z}$ contributing to $Y_1$.
In parentheses, we report the standard deviation of the coefficients over 3 training runs over different subsets of data.
}\label{tab:brain_wbfm_surrogate}
\begin{tabular}{|l|c|c|c|c|c|}
\hline
   Variable &          Brain Stem &                CSF &        Gray Matter &        Hippocampus &          Ventricles \\
\hline
      $Y_1$ & $-0.03\ (\pm 0.01)$ & $0.01\ (\pm 0.00)$ & $0.04\ (\pm 0.00)$ & $0.05\ (\pm 0.00)$ & $-0.02\ (\pm 0.00)$ \\
$\psi_1(\mathbf{Z})$ & $-0.20\ (\pm 0.07)$ & $0.06\ (\pm 0.22)$ & $0.25\ (\pm 0.04)$ & $0.22\ (\pm 0.05)$ & $-0.17\ (\pm 0.08)$ \\
\hline
\end{tabular}
\end{table}

\section{Discussion}
In this work, we approached in a systematic manner the task of recovering the latent signal shared between scalar variables, by formalizing the problem setting and  defining an evaluation procedure for model benchmarking, and proposed a new method, \ac{DCID}, for solving the task.

\subsection{Results Summary}
By conducting experiments in controlled settings on synthetic data we could analyze model performance wrt. properties of the latent variables.
Notably, we observed that the baseline models performed poorly when the shared variables were not strongly dominating the signal in the data, which is arguably the more realistic setting.
The \ac{DCID} model proved more robust in these scenarios, outperforming the baselines in most cases.
We note, however, that it was still sensitive to the magnitude of the shared signal, and, interestingly, exhibited a loss of performance when the shared signal was strongly dominating.
Investigating the loss of performance in the strong-signal regime, and improving robustness in the low-signal one are thus two natural directions for future work.
Nevertheless, we believe that \ac{DCID} can serve as an easy-to-implement, yet effective baseline.
\subsection{Limitations}
A main assumption of the method is that the observed variables $\mathbf{X}$ are rich in information, preserving the signal about the latent variables.
Since, in practice, we do not observe the latent variables, we cannot test whether this assumption holds.
As a substitute safety measure, we can assess the performance in predicting the observed target variables $Y$.
If these cannot be predicted accurately, then it is unlikely that the model will correctly recover the latent variables either.
Furthermore, we note that the method should not be mistaken as allowing to reason about causal relations between variables.
It could, however, be used as part of a preprocessing pipeline in a causal inference setting, e.g., for producing candidate variables for mediation analysis.
\section{Acknowledgements}
This research was funded by the HPI research school on Data Science and Engineering.
Data used in the preparation of this article were obtained from the UK Biobank Resource under Application Number 40502.
\section{Ethical Considerations}
As mentioned in the main text (Section~\ref{sssec:exp_settings_mri}), we conducted the brain \ac{MRI} experiments on the ``white-British'' subset of the \ac{UKB} dataset.
This was done to avoid unnecessary confounding, as the experiments were meant as a proof of concept, rather than a strict medical study. 
When conducting the latter, measures should be taken to include all available ethnicities whenever possible, in order to avoid increasing the already existing disparities in representations of ethnic minorities in medical studies.

\bibliography{bibliography}

\end{document}